\documentclass[runningheads]{llncs}

\usepackage{graphicx}
\usepackage{xcolor}
\usepackage{amssymb}
\usepackage{amsmath}
\usepackage[hyphens]{url}
\usepackage{multirow}

\newcommand{\ii}{\boldsymbol{i}}
\newcommand{\jj}{\boldsymbol{j}}
\newcommand{\kk}{\boldsymbol{k}}
\newcommand{\quat}[1]{{#1}_0 + {#1}_1 \ii + {#1}_2 \jj + {#1}_3 \kk}

\begin{document}

\title{Quaternion-Valued Convolutional Neural Network Applied for Acute Lymphoblastic Leukemia Diagnosis
\thanks{This work was supported in part by Coordena\c{c}\~ao  de Aperfei\c{c}oamento de Pessoal de N\'ivel Superior - Brasil (CAPES) - Finance Code 001.}
}
\titlerunning{Quaternion-Valued CNNs for Acute Lymphobastic Leukemian Diagnosis}
%
\author{Marco Aurélio Granero\inst{1}\orcidID{0000-0002-1993-1898} \and Cristhian Xavier Hernández\inst{2}\orcidID{0000-0001-8530-9880} \and Marcos Eduardo Valle\inst{3}\orcidID{0000-0003-4026-5110}}
\authorrunning{Granero, M. A. et al.}
\institute{Inst. Federal de Educação, Ciência e Tecnologia de São Paulo, São Paulo, Brazil. \and 
Escuela Superior Politécnica del Litoral, Guayaquil, Ecuador. \and Universidade Estadual de Campinas, Campinas, Brazil.
\newline \email{granero@ifsp.edu.br and valle@ime.unicamp.br}
}
\maketitle              

\begin{abstract}
The field of neural networks has seen significant advances in recent years with the development of deep and convolutional neural networks. Although many of the current works address real-valued models, recent studies reveal that neural networks with hypercomplex-valued parameters can better capture, generalize, and represent the complexity of multidimensional data. This paper explores the quaternion-valued convolutional neural network application for a pattern recognition task from medicine, namely, the diagnosis of acute lymphoblastic leukemia. Precisely, we compare the performance of real-valued and quaternion-valued convolutional neural networks to classify lymphoblasts from the peripheral blood smear microscopic images. The quaternion-valued convolutional neural network achieved better or similar performance than its corresponding real-valued network but using only 34\% of its parameters. This result confirms that quaternion algebra allows capturing and extracting information from a color image with fewer parameters.

\end{abstract}

\section{Introduction}

In recent years, machine learning has influenced how we solve a variety of real-world problems. Indeed, artificial neural networks (NN) outperformed many state-of-the-art approaches in several applications with the development of deep neural networks (DNN) and convolutional neural networks (CNN) architectures.

Most neural network architectures are real-valued neural networks (RVNN). In such architectures, the input data is arranged into real-valued vectors, matrices, or tensors to be processed by the neural network. In some sense, this approach assumes that all the input data components have equal importance and, thus, they are all evaluated in the same way. However, in some cases, the data sets contain multidimensional information that requires a specific approach to treat them as single entities. For example, a pixel's color is obtained by combining the red, green, and blue components in image processing. The three coordinates position in the color space represents a plethora of colors such as pink or brown, and the color information is lost if the components are treated separately \cite{Parcollet2020ANetworks}. In some practical image recognition tasks, the complexity of the color space needs to be captured by the neural networks to generalize well and represent the multidimensional nature of the colors \cite{kusamichi04,isokawa2009}. Indeed, Parcollet et al. showed that RVNNs might fail to capture the color information  \cite{parcollet2018}. Also, Matsui et al. remarked that RVNNs are not able to preserve the 3D shape of an input object when transformed into the 3D space \cite{matsui04}. From these remarks, neural networks based on hypercomplex numbers, such as complex and quaternions, have been proposed and extensively investigated in the last years. 

\subsection{Complex and Quaternion-Valued Neural Networks}

A complex-valued neural network (CVNN) is based on the algebra of complex numbers, which allows preserving or treating the relationship between magnitude and the phase information during the learning \cite{Parcollet2020ANetworks}. Furthermore, the algebraic structure of complex numbers yields  CVNNs better generalization capability \cite{hiroseyoshida2012} besides being easier to train \cite{nitta2002}. As long as the processed information are correlated two-dimensional data, CVNNs mostly outperformed or at least matched the real-valued ones \cite{aizenberg2011,aizenberg18wcci,hirose12,mandic2009,Trabelsi17complex}. 

The encouraging performance of CVNNs inspired the development of qua\-ter\-ni\-on-valued neural networks (QVNNs). QVNNs use quaternion algebra and can represent colors efficiently, with the advantage of fully representing colors through unique structures \cite{arena97conflito}.

As far as we know, the first QVNN has been introduced by Arena et al. \cite{arena97conflito}, who developed a specific backpropagation algorithm able to learn the local relations that exist between quaternions. Furthermore, like the real-valued neural networks, single hidden layer QVNNs are universal approximators \cite{arena97conflito}.  An extensive list of applications and investigations with different QVNN architectures can be found in references \cite{bayro2018,gaudet-maida2018,ogawa2016,onyekpe2021,Parcollet2020ANetworks,Parcollet2018b,takahashietal2017}. A detailed up-to-date review on quaternion-valued neural networks, including some of their successful applications, can be found at \cite{Parcollet2020ANetworks}.

In contrast to RVNN, which represented color channels as independent variables, QVNN can benefit from representing colors as single quaternions. For example, Greenblatt et al. applied a QVNN model to prostate cancer \cite{greenblattetal2013}. Gaudet and Maida investigated the use of quaternion-valued convolutional neural networks (QVCNN) for image processing \cite{gaudet-maida2018}. Pavllo et al. modeled human motion using QVNNs \cite{Pavllo_2019}. Zhu et al. proposed a QVCNN for color image classification and denoising tasks \cite{zhu2019quaternion}. The localization of color image splicing by a fully quaternion-valued convolutional network was explored by Chen et al. \cite{chen_etal2019}. A deformable quaternion Gabor convolutional neural network for recognition of color facial expression was proposed by Jin et al. \cite{jinetal2020}. Takahashi et al. have merged histograms of oriented gradients (HOG) for human detection with a QVNN to determine human facial expression \cite{takahashi_etal2014}. Quaternion multi-layer perceptron has been successfully applied to polarimetric synthetic aperture radar (PolSAR) land classification \cite{kinugawa18,shang14}.

\subsection{Contributions and the Organization of the Paper}

\begin{figure}[t]
    \begin{center}
        \includegraphics[scale=0.5]{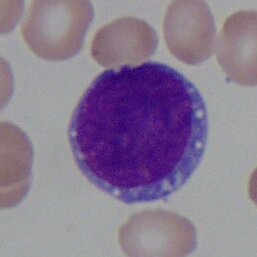}
    \end{center}
    \vspace{-0.25cm}
    \caption{Candidate cell to be a lymphoblast, from ALL-IDB dataset \cite{scotti_ALL_IDB}.} \label{img_cell}
\end{figure}

Corroborating with the development of hypercomplex-valued neural networks, we present a quaternion-valued convolutional neural network (QVCNN) development to classify isolated white cells as lymphoblasts.  Precisely, the QVCNN receives a white cell image like the one shown in Figure \ref{img_cell} and classifies it as a lymphoblast or not. 
The classification of lymphoblast is essential for diagnosing acute lymphoblastic leukemia, a kind of blood cancer. The performance of the QVCNN is compared with a real-valued convolutional neural network with a similar architecture.

The paper is structured as follows: Section \ref{sec:ALL} presents the medical problem of acute lymphoblastic leukemia and presents a literature review on the computer-aided diagnosis of leukemia. Section \ref{sec:CNNs} addresses real-valued and quaternion-valued convolutional neural networks. The experimental results are detailed in section \ref{sec:experiments}. The section \ref{sec: ConRemarksANDfutureWorks} presents the concluding remarks and future works.

\section{Acute Lymphoblastic Leukemia (ALL)} \label{sec:ALL}

According to the national cancer institute of the United States, acute lymphoblastic leukemia (ALL) is a type of leukemia, cancer in the blood, that appears and multiplies rapidly \cite{def}. ALL is characterized by the presence of many lymphoblasts in the blood and also in the bone marrow. In this context, a lymphoblast is an immature cell that can be converted into a mature lymphocyte \cite{def_linf}. 

\begin{figure}[t]
    \begin{center}
        \includegraphics[width=\columnwidth]{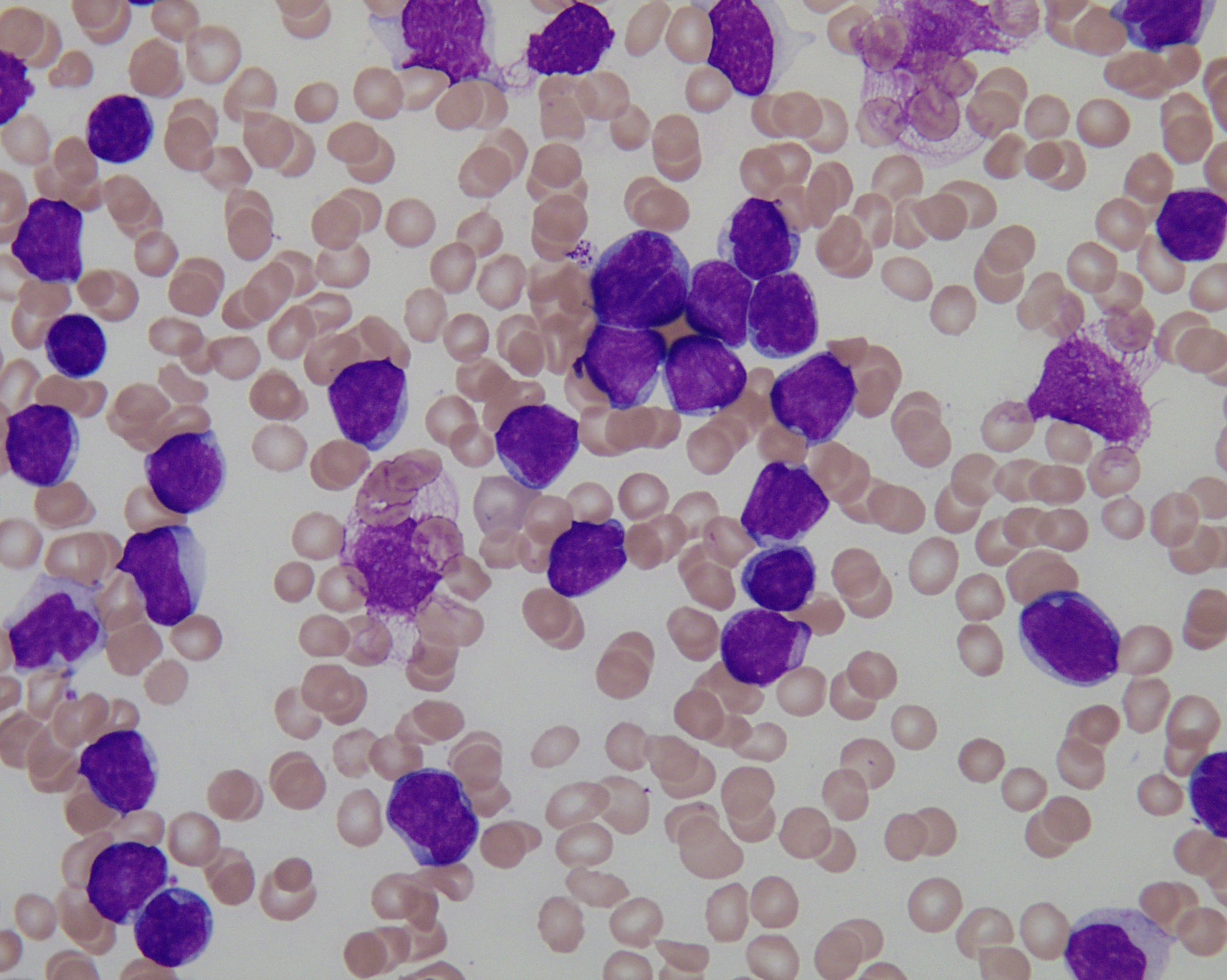}
    \end{center}
    \vspace{-0.25cm}
    \caption{Blood smear image from ALL-IDB dataset \cite{scotti_ALL_IDB}.} \label{img_blood_smear}
\end{figure}

There are several methods used for the diagnosis of ALL that can be found in the literature \cite{metodos_diag}, including the peripheral blood smear technique \cite{frotis_sangre}. The peripheral blood smear technique allows observing the information of a blood sample taken from the patient through a microscope. A specialist (hematologist) counts the number of lymphoblasts observed by microscope and, based on that, makes a diagnosis \cite{art_diag}. Figure \ref{img_blood_smear} shows a picture of a blood smear that a hematologist sees for analysis. It is worth mentioning that the white cells appear stained with a bluish-purple coloration, which serves as a guide to find lymphoblasts.

The manual counting of lymphoblasts under the microscope is a somewhat dull task that takes much time from a professional who could be more productive in other matters. In effect, the time spent analyzing the microscope image has an economical cost because a specialist has significant value in the labor market. In addition, the analysis can be affected by human factors such as tiredness and stress. The operator's experience also plays an important role, and therefore, there is a subjectivity component affecting the results of the lymphoblast count. For these reasons, computational models to perform automatic lymphoblast counting in a blood smear image have been proposed in the literature \cite{art_historia}. 

Many methods divide the problem of automatic lymphoblast counting into two stages. The first stage, usually called the identification phase, aims to find white cells to be lymphoblast. Labeling a candidate cell as a lymphoblast or healthy cell is performed in the second stage, referred to as the classification phase. In this paper, we use real-valued and quaternion-valued convolutional neural networks to classify white cells, that is, in the second stage of the blood smear image analysis. In the following sections, we review real-valued and quaternion-valued neural networks. Before, however, we provide a literature review on automatic leukemia diagnosis methods.

\subsection{Computer-Aided Diagnosis of Leukemia: Literature Review} \label{sec:CAD}

Current literature has shown a large number of studies on computer-aided leu\-ke\-mia diagnosis with different approaches, including support vector machines (SVM), k-nearest neighbor (k-NN), principal component analysis (PCA), naive Bayes classifier, and random forest \cite{IoMT_2020}.

In \cite{ref_32}, the authors used 60 sample images to develop a model to detected ALL using kNN and naive Bayes classifier with 92.8\% accuracy. A method to extract features of microscopic images using discrete orthogonal Stockwell transform (DOST) and linear discriminant analysis (LDA) has been proposed in \cite{MISHRA2019}. The paper \cite{VOGADO2018} applies three pre-trained CNN architectures to extract features for image classification. In \cite{Ahmed_2019}, a CNN reached 88.25\% of accuracy in classifying ALL versus healthy cells. To distinguish between the four subtypes of leukemia, this CNN hits 81.74\% accuracy. Using ALL-IDB dataset, \cite{Acharya_2019} presents a k-medoids algorithm with 98.60\% accuracy to classify white blood cells. Furthermore, a method based on generative adversarial optimization (GAO) \cite{Tuba2019}, a neural network with statistic features \cite{Aljaboriy2019}, and a deep CNN with chronological sine-cosine algorithm (SCA) \cite{Jha2019} have been proposed for ALL detection with 93.84\%, 97.07\%, and 98.70\% accuracy, respectively.

A table summarizing the results from 16 papers on automated detection of leukemia and its subtypes can be found in \cite{IoMT_2020}. This reference also presents a framework for automated leukemia diagnosis based on the ResNet-34 \cite{He2016_Resnet} and the DenseNet-121 \cite{Huang2017_DenseNet}. The accuracy reported was 99.56\% for the ResNet-34 and 99.91\% for the DenseNet-121 \cite{IoMT_2020}.

\section{Convolutional Neural Networks} \label{sec:CNNs}

In many machine learning applications, identifying appropriate representations of a large amount of data is usually challenging. A successful model must efficiently encode local relations within the input resources and their structural relations. Moreover, an adequate representation of data also offers a positive side effect by reducing the number of neural parameters needed to well-learn the input features, leading to a natural solution to the overfitting phenomenon \cite{Parcollet2020ANetworks}.

Convolutional neural networks (CNN) are feed-forward neural networks with a robust feature representation method widely applied in machine learning. For example, the ResNet set a milestone in 2015 by outperforming humans in the ImageNet competition \cite{He2016_Resnet,Garcia-Retuerta2020}.  The successful AlexNet  \cite{Krizhevsky2017_Alexnet} also inspired the development of many novel CNNs including the VGG \cite{simonyan2015_VGG} and the DenseNet  \cite{Huang2017_DenseNet}. In addition, deep neural networks have been successfully used, for example, for segmentation tasks as well as for the automatic classification of objects in images \cite{he2018mask,Hu2018_squeeze}.

One crucial aspect of the deep networks is the convolution layer, which extracts features from high-dimensional data through a set of convolution kernels \cite{zhu2019quaternion}. Although convolutions perform well in many practical situations, it has some drawbacks in color image processing tasks. Firstly, a convolution layer sums up the outputs corresponding to different channels and ignores their complicated interrelationships. As a consequence, it may eventually lose important information of a color image. Secondly, simply summing up the outputs gives too many degrees of freedom, and thus, the network has a high risk of overfitting even when imposing heavy regularization terms \cite{zhu2019quaternion}. 
Accordingly, García-Retuerta et al. argue that quaternion-valued neural networks may have a significant advantage in color image processing tasks because of quaternion's four-dimensional algebraic structure \cite{Garcia-Retuerta2020}. The following section reviews the basic concepts of quaternion-valued convolutional neural networks.

\subsection{Quaternion-valued Convolutional Neural Networks}

Quaternions are a four-dimensional extension of complex numbers. Developed by Hamilton in 1843, the set of all quaternions is defined by
\begin{equation}
     \mathbb{H} = \{q = \quat{q}: q_0,q_1,q_2,q_3 \in  \mathbb{R} \}
\end{equation}
where $q_0$ is the real part of a quaternion, $q_1$, $q_2$, and $q_3$ denote the imaginary components while $\ii$, $\jj$, $\kk$ are the hypercomplex units. The product of the hypercomplex units is governed by the following identities, knows as Hamilton rules:
\begin{equation}
    \begin{array}{c}
        \ii^2 = \jj^2 = \kk^2 = \ii \jj \kk = -1. 
    \end{array}
\end{equation}
Alternatively, a quaternion can be written as 
\begin{equation} \label{eq:quat2complex}
q = (q_0+q_1\ii) + (q_2+q_3\ii)\jj = z_0 + z_1 \jj,
\end{equation}
where $z_0 = q_0 + q_1\ii$ and $z_1 = q_2+q_3\jj$ are complex numbers.

The addition of quaternions is performed adding the real and imaginary components. Precisely, given $p = \quat{p}$ and $q =\quat{q}$, their sum is 
\begin{equation}
    p+q = (p_0+q_0) + (p_1+q_1)\ii + (p_2+q_2)\jj + (p_3+q_3)\kk.
\end{equation}
The main result in quaternion algebra is the Hamilton product between two quaternions $p = \quat{p}$ and $q = \quat{q}$, denoted by $p \otimes q$ and defined by 
\begin{equation} \label{hamiltonproduct}
    \begin{array}{rl}
    p \otimes q &= (p_0 q_0 - p_1 q_1 - p_2 q_2 - p_3 q_3) 
    + \, (p_0 \, q_1 + p_1 \, q_0 + p_2 \, q_3 - p_3 \, q_2) \, \ii \\
    & + \, (p_0 \, q_2 - p_1 \, q_3 + p_2 \, q_0 + p_3 \, q_1) \, \jj
    + \, (p_0 \, q_3 + p_1 \, q_2 - p_2 \, q_1 + p_3 \, q_0) \, \kk 
    \end{array}
\end{equation}
Quaternions and quaternion algebra allow building processing entities composed of four elements that share information via the Hamilton product. 

According to Gaudet and Maida \cite{gaudet-maida2018}, a quaternion-valued convolutional layer is obtained convolving a quaternion-valued filter matrix 
$\boldsymbol{W} = \quat{\boldsymbol{W}}$
by a quaternion-valued vector $\boldsymbol{h} = \quat{\boldsymbol{h}}$. Here, $\boldsymbol{W}_0$, $\boldsymbol{W}_1$, $\boldsymbol{W}_2$, and $\boldsymbol{W}_3$ are real-valued matrices while $\boldsymbol{h}_0$, $\boldsymbol{h}_1$, $\boldsymbol{h}_2$, and $\boldsymbol{h}_3$ are real-valued vectors. Details on the implementation of quaternion-valued convolutional layers can be found in \cite{gaudet-maida2018}.

\section{Computational Experiments}  \label{sec:experiments}

Let us compare real-valued and quaternion-valued convolutional neural networks' performance for classifying a white cell image as a lymphoblast. Both real-valued and quaternion-valued neural networks have been implemented in python using the Keras and Tensorflow libraries.

\begin{figure}[!ht]
    \centering
    \includegraphics[width=\columnwidth]{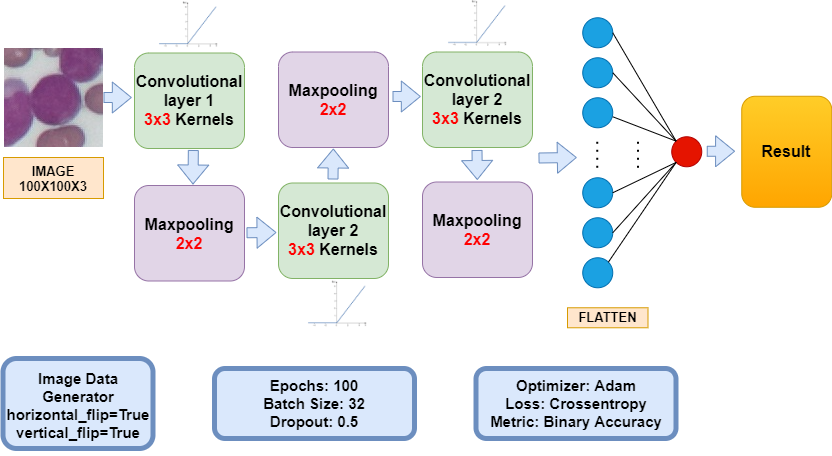}
    \caption{RVCNN and QVCNN architectures.}
    \label{fig:modelo_redeneural}
\end{figure}

The real-valued model is a sequential feed-forward network composed of three convolutional layers, three max-pooling layers, and a dense layer. Precisely, the first convolutional layer has $32$ filters with a $(3,3)$ kernel and ReLU activation function. A max-pooling follows the convolutional layer with a $(2,2)$ kernel. The second and third two-dimensional convolutional layers have $64$ and $128$ filters and have ReLU activation functions. Furthermore, they are also followed by max-pooling layers with $(2,2)$ kernels. Figure \ref{fig:modelo_redeneural} shows the architecture of the real-valued convolutional neural network. The total number of trainable parameters of the real-valued convolutional neural network is $106,049$.

The quaternion-valued convolutional neural network has been designed similarly. Precisely, to maintain the same parameter budget among the real and quaternion-valued models, the number of filters per layer of the real-valued network was divided by four to built a quaternion-valued convolution. Thus, the quaternion-valued convolutional neural network has the same structure as the real-valued network depicted in Figure \ref{fig:modelo_redeneural}, but with a quarter of the number of filters per layer. The number of trainable parameters of the quaternion-valued CNN model is $36,353$. Table \ref{table1} summarizes the number of trainable parameters of both neural networks per layer.

\begin{table}[!ht]
\centering
\begin{tabular}{|c|c|c|c|c|c|c|c|c|} 
\cline{2-9}
\multicolumn{1}{l|}{} & \multicolumn{2}{c|}{Conv Layer 1} & \multicolumn{2}{c|}{Conv Layer 2} & \multicolumn{2}{c|}{Conv Layer 3} & \multicolumn{2}{c|}{Dense Layer}    \\ 
\hline
                      & RVCNN & QVCNN           & RVCNN   & QVCNN         & RVCNN   & QVCNN         & RVCNN   & QVCNN           \\ 
\hline
Filters (3,3)         & 32   & 8                    & 64     & 16                 & 128    & 32                 & 1      & 1                    \\ 
\hline
Activation           & \multicolumn{2}{c|}{ReLU}   & \multicolumn{2}{c|}{ReLU}   & \multicolumn{2}{c|}{ReLU}   & \multicolumn{2}{c|}{None}  \\ 
\hline
Max pooling           & \multicolumn{2}{c|}{(2,2)}  & \multicolumn{2}{c|}{(2,2)}  & \multicolumn{2}{c|}{(2,2)}  & \multicolumn{2}{c|}{-}        \\ 
\hline
Parameters            & 896  & 320                  & 18,496 & 4,672              & 73,856 & 18,560             & 12,801 & 12,801               \\
\hline
\end{tabular}
    \caption{Parameters of real and quaternion-valued neural networks}
    \label{table1}
\end{table}

The dense layer of both real-valued and quaternion-valued networks has a single output neuron without activation function. Such a single neuron is used to classify the input image as a lymphoblast or not. Moreover, the parameters of all layers have been initialized according to Glorot and Bengio \cite{glorot2010}. The optimizer used was $Adam$, an algorithm based on the stochastic gradient descent method with adaptive estimation of first-order and second-order moments.

To evaluate the performance of the RVCNN and QVCNN classifiers, we used the \textit{ALL-IDB: The Acute Lymphoblastic Leukemia Image Database for Image Processing}  provided by the ``Università Degli Studi di Milano'' \cite{scotti_ALL_IDB}. This image database contains 260 images of white blood cells with $257 \times 257$ pixels, labeled by experts and evenly distributed among lymphoblast and health cells. Figure \ref{img_cell} shows an example of a color image used in the computational experiment.

We resized the $257 \times 257$ white blood cells images to $100 \times 100$ pixels. Also, the set of 260 color images was randomly divided into training and test images with different ratios. Data augmentation has been applied on the training set to improve the accuracy of the convolutional neural networks. Precisely, the images used for training were all submitted to a pre-processing data generation, which consists of obtaining new images through horizontal and vertical flips.

In our experiments, images were converted to RGB (red, green, and blue) and HSV (hue, saturation, and value) color spaces and used as input to neural networks. As a consequence, we performed the four experiments detailed in Table \ref{table2}. The first experiment considers a real-valued CNN whose input is obtained by concatenating the three RGB channels in a single tensor with values in the unit interval $[0, 1]$. The second experiment also considers real-valued CNNs, but the input is obtained by concatenating the three HSV channels. Here, hue is arranged in a radial slice $H \in [0,2\pi)$ while saturation and value belong to the unit interval, i.e., $S, \ V \in [0,1]$. 

\begin{table}[!ht]
    \centering
    \begin{tabular}{|c|c|c|} 
        \hline
        \quad \ Neural Network \quad \ & \ \quad Input \quad \ &  Input Structure              \\ 
        \hline
        Real-valued CNN  & RGB & Concateneted channel  \\ 
        \hline
        Real-valued CNN  & HSV & Concateneted channel  \\ 
        \hline
        Quaternion-values CNN  & RGB & Quaternions  encoded using \eqref{eq:quaternion_RGB}  \\ \hline
        Quaternion-valued CNN  & HSV & \quad \ Quaternions encoded using \eqref{eq:quaternion_HSV} \quad \ \\
        \hline
    \end{tabular}
    \caption{Experiments with real and quaternion-valued neural networks}
    \label{table2}
\end{table}

The last two experiments were performed using quaternion-valued CNN. Specifically, in the third experiment, the RGB image is encoded in a quaternion structure with real part null, and each channel as one imaginary part of a quaternion as follows: 
\begin{equation} \label{eq:quaternion_RGB}
    q = 0 + R \ \ii + G \ \jj + B \ \kk.
\end{equation}
Finally, in the fourth experiment, a color is encoded in a quaternion through the following expression using the HSV representation:
\begin{equation} \label{eq:quaternion_HSV}
q = S \cos(H) + S \sin(H) \ \ii + V \cos(H) \ \jj + V \sin(H)  \ \kk.
\end{equation}

\begin{figure}[!ht]
    \centering
    \includegraphics[width=\columnwidth]{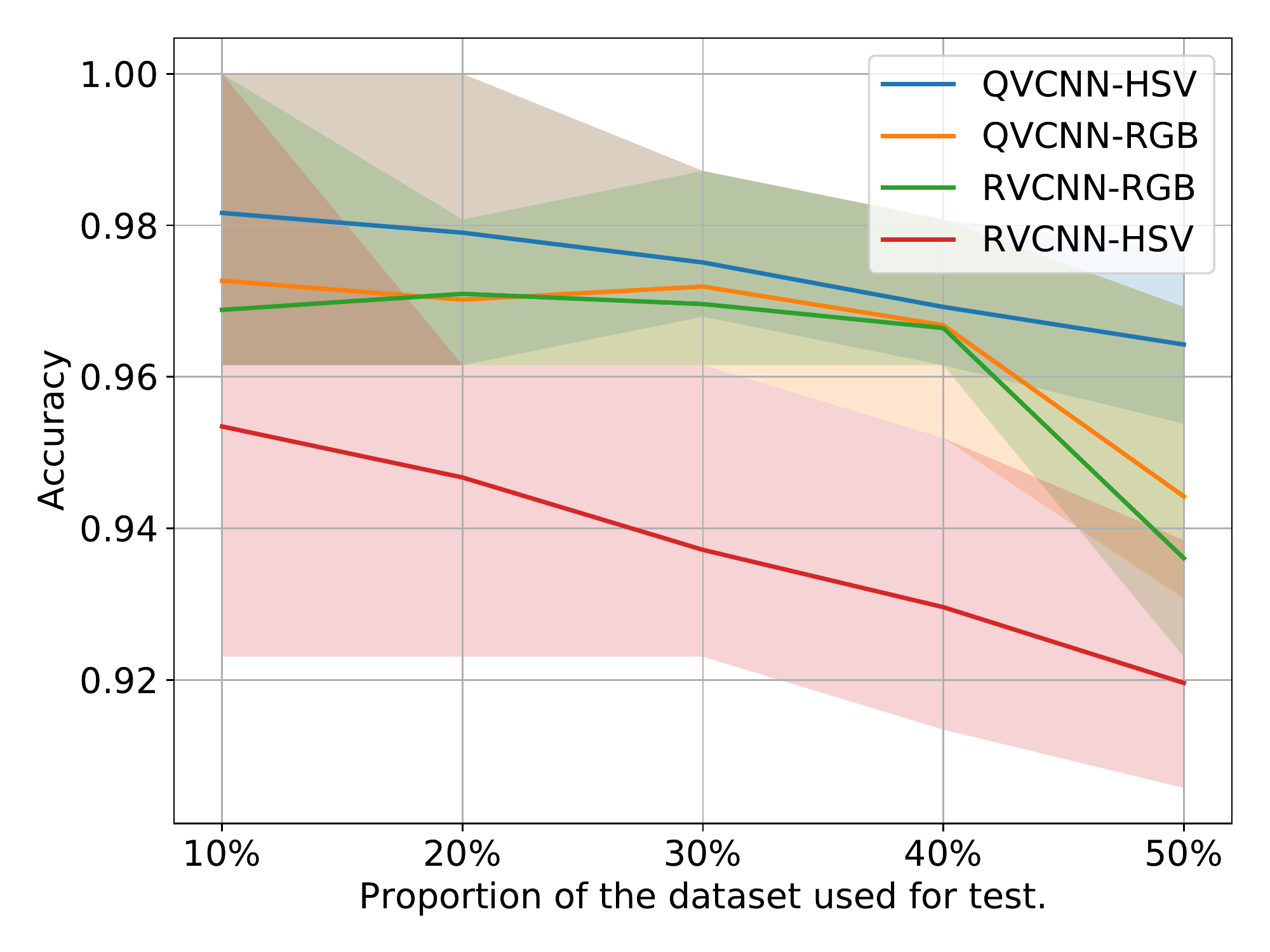} 
    \caption{Accuracy by the percentage of the dataset used for testing.} \label{fig:test_size}
\end{figure}

The dataset has been divided into training and test sets with 5 different training/test ratios and trained by 100 epochs. One hundred simulations were performed for each different training/test ratio and, the average and standard deviation of the accuracy was calculated. Figure \ref{fig:test_size} presents the average accuracy of both real-valued and quaternion-valued convolutional neural networks for different percentages used for testing the networks in the four experiments. This figure also presents the interval between the $25\%$ and the $75\%$ quantiles of accuracy as shaded area.

Note from Figure \ref{fig:test_size} that the quaternion-valued convolutional neural network with images in HSV color space (QVCNN-HSV) obtained the best performance, reaching 98.2\% of accuracy in the test phase with 10\% of training/test ratio. 

The real and quaternion-valued networks with RGB encoded images exhibited similar performance, with accuracy between $[93.6\%, \ 97.1\%]$ and $[94.4\%,$ $97.3\%]$, respectively, depending on the ratio training/test. The real and quaternion-valued CNN models with RGB encoded images exhibited statistically equivalent performances. The real-valued neural network with HSV encoded images yielded the worst performance, reaching an average accuracy of 95.3\% in the best case.

Concluding, the QVCNN-HSV exhibits a better generalization capability than the QVCNN-RGB, RVCNN-RGB, and RVCNN-HSV models. Moreover, the performance of the quaternion-valued convolutional neural network with images encoded using the HSV color space and \eqref{eq:quaternion_HSV} compares well with the results reported in the literature (see Section \ref{sec:CAD}). However, the quaternion-valued convolutional neural network is much simpler than many of the architectures considered previously. 

\section{Concluding Remarks and Future Works} \label{sec: ConRemarksANDfutureWorks}

Acute lymphoblastic leukemia is characterized by many lymphoblasts in the blood and the bone marrow. Such disease can be diagnosticated by counting the number of lymphoblasts in a blood smear microscope image. This paper investigated the application of convolutional neural networks for classifying a white cell as lymphoblast or not. Precisely, we compared the performance of real-valued and quaternion-valued models. The QVCNN with input images encoded using the HSV color space showed the best result in our experiments. Also, the performance of the QVCNN is comparable with other deeper neural networks from the literature, including the ResNet and the DenseNet \cite{IoMT_2020}. This computational experiment suggests that quaternion-valued neural networks exhibit better generalization capability than the real-valued convolutional neural network, possibly because it treats colors as single quaternion entities. Furthermore, it is noticeable that the quaternion-valued convolutional neural network has about 34\% of the parameters of the corresponding real-valued model.

We plan to develop neural networks that segment and classify white blood cells on a blood smear microscope image as future work.  Further research can also address the application of QVCNN for the classification of other types of leukemia.


\begin{thebibliography}{10}
\providecommand{\url}[1]{\texttt{#1}}
\providecommand{\urlprefix}{URL }
\providecommand{\doi}[1]{https://doi.org/#1}

\bibitem{Acharya_2019}
Acharya, V., Kumar, P.: Detection of acute lymphoblastic leukemia using image
  segmentation and data mining algorithms. Medical \& Biological Engineering \&
  Computing  (2019). \doi{https://doi.org/10.1007/s11517-019-01984-1}

\bibitem{Ahmed_2019}
Ahmed, N., Yigit, A., Isik, Z., Alpkocak, A.: Identification of leukemia
  subtypes from microscopic images using convolutional neural network.
  Diagnostics (Basel)  \textbf{9}(3) (2019). \doi{10.3390/diagnostics9030104}

\bibitem{aizenberg2011}
Aizenberg, I., Alexander, S., Jackson, J.: {Recognition of blurred images using
  multilayer neural network based on multi-valued neurons}. 2011 41st IEEE
  International symposium on multiple-valued logic pp. 282--287 (2011)

\bibitem{aizenberg18wcci}
Aizenberg, I., Gonzalez, A.: {Image Recognition using MLMVN and Frequency
  Domain Features}. In: 2018 International Joint Conference on Neural Networks
  (IJCNN). pp.~1--8 (2018). \doi{10.1109/IJCNN.2018.8489301}

\bibitem{Aljaboriy2019}
Aljaboriy, S., Sjarif, N., Chuprat, S., Abduallah, W.: Acute lymphoblastic
  leukemia segmentation using local pixel information. Pattern Recognition
  Letters  \textbf{125},  85--90 (03 2019). \doi{10.1016/j.patrec.2019.03.024}

\bibitem{arena97conflito}
Arena, P., Fortuna, L., Muscato, G., Xibilia, M.G.: {Multilayer Perceptrons to
  Approximate Quaternion Valued Functions}. Neural Networks  \textbf{10}(2),
  335--342 (1997). \doi{10.1016/S0893-6080(96)00048-2}

\bibitem{bayro2018}
Bayro-Corrochano, E., Lechuga-Gutiérrez, L., Garza-Burgos, M.: {Geometric
  techniques for robotics and hmi: Interpolation and haptics in conformal
  geometric algebra and control using quaternion spike neural networks}. Robot
  Auton System  \textbf{104},  72--84 (2018)

\bibitem{IoMT_2020}
Bibi, N., Sikandar, M., Din, I.U., Almogren, A., Ali, S.: Iomt-based automated
  detection and classification of leukemia using deep learning. Journal of
  Healthcare Engineering  (2020). \doi{10.1155/2020/6648574}

\bibitem{chen_etal2019}
Chen, B., Gao, Y., Xu, L., Hong, X., Zheng, Y., Shi, Y.Q.: {Color image
  splicing localization algorithm by quaternion fully convolutional networks
  and superpixel-enhanced pairwise conditional random field[J]}. Mathematical
  Biosciences and Engineering  \textbf{6}(16),  6907--6922 (2019).
  \doi{10.3934/mbe.2019346}

\bibitem{Garcia-Retuerta2020}
García-Retuerta, D., Casado-Vara, R., Martin-del Rey, A., De~la Prieta, F.,
  Prieto, J., Corchado, J.M.: Quaternion neural networks: State-of-the-art and
  research challenges. In: Analide, C., Novais, P., Camacho, D., Yin, H. (eds.)
  Intelligent Data Engineering and Automated Learning -- IDEAL 2020. pp.
  456--467. Springer International Publishing, Cham (2020)

\bibitem{gaudet-maida2018}
Gaudet, C.J.;~Maida, A.: {Deep quaternion networks}. International joint
  conference on neural networks (IJCNN) pp.~1--8 (2018)

\bibitem{glorot2010}
Glorot, X., Bengio, Y.: Understanding the difficulty of training deep
  feedforward neural networks. In: Teh, Y.W., Titterington, M. (eds.)
  Proceedings of the Thirteenth International Conference on Artificial
  Intelligence and Statistics. Proceedings of Machine Learning Research,
  vol.~9, pp. 249--256. PMLR, Chia Laguna Resort, Sardinia, Italy (13--15 May
  2010), \url{http://proceedings.mlr.press/v9/glorot10a.html}

\bibitem{greenblattetal2013}
Greenblatt, A., Mosquera-Lopez, C., Agaian, S.: Quaternion neural networks
  applied to prostate cancer gleason grading. In: 2013 IEEE International
  Conference on Systems, Man, and Cybernetics. pp. 1144--1149 (2013).
  \doi{10.1109/SMC.2013.199}

\bibitem{he2018mask}
He, K., Gkioxari, G., Dollár, P., Girshick, R.: Mask r-cnn (2018)

\bibitem{He2016_Resnet}
He, K., Zhang, X., Ren, S., Sun, J.: Deep residual learning for image
  recognition. In: 2016 IEEE Conference on Computer Vision and Pattern
  Recognition (CVPR). pp. 770--778 (2016). \doi{10.1109/CVPR.2016.90}

\bibitem{hirose12}
Hirose, A.: {Complex-Valued Neural Networks}. Studies in Computational
  Intelligence, Springer, Heidelberg, Germany, 2nd editio edn. (2012)

\bibitem{hiroseyoshida2012}
Hirose, A., Yoshida, S.: Generalization characteristics of complex-valued
  feedforward neural networks in relation to signal coherence. IEEE
  Transactions on Neural Networks and Learning Systems  \textbf{23}(4),
  541--551 (2012). \doi{10.1109/TNNLS.2012.2183613}

\bibitem{Hu2018_squeeze}
Hu, J., Shen, L., Sun, G.: Squeeze-and-excitation networks. In: 2018 IEEE/CVF
  Conference on Computer Vision and Pattern Recognition. pp. 7132--7141 (2018).
  \doi{10.1109/CVPR.2018.00745}

\bibitem{Huang2017_DenseNet}
Huang, G., Liu, Z., Van Der~Maaten, L., Weinberger, K.Q.: Densely connected
  convolutional networks. In: 2017 IEEE Conference on Computer Vision and
  Pattern Recognition (CVPR). pp. 2261--2269 (2017).
  \doi{10.1109/CVPR.2017.243}

\bibitem{isokawa2009}
Isokawa, T., Matsui, N., Nishimura, H.: {Quaternionic Neural Networks:
  Fundamental Properties and Applications}. In: Complex-valued neural networks:
  utilizing high-dimensional parameters. pp. 411--439 (2009)

\bibitem{Jha2019}
Jha, K.K., Sekhar~Dutta, H.: Mutual information based hybrid model and deep
  learning for acute lymphocytic leukaemia detection in single cell blood smear
  images. Computer Methods and Programs in Biomedicine  \textbf{179},  104987
  (07 2019). \doi{10.1016/j.cmpb.2019.104987}

\bibitem{jinetal2020}
Jin, L., Zhou, Y., Liu, H., Song, E.: Deformable quaternion gabor convolutional
  neural network for color facial expression recognition. In: 2020 IEEE
  International Conference on Image Processing (ICIP). pp. 1696--1700 (2020).
  \doi{10.1109/ICIP40778.2020.9191349}

\bibitem{frotis_sangre}
Kasvi: Hematologia: Como é realizada a técnica de esfregaço de sangue?
  \url{https://kasvi.com.br/esfregaco-de-sangue-hematologia/} (2021), accessed: 
  2021-02-18

\bibitem{kinugawa18}
Kinugawa, K., Shang, F., Usami, N., Hirose, A.: {Isotropization of
  Quaternion-Neural-Network-Based PolSAR Adaptive Land Classification in
  Poincare-Sphere Parameter Space}. IEEE Geoscience and Remote Sensing Letters
  \textbf{15}(8),  1234--1238 (2018). \doi{10.1109/LGRS.2018.2831215}

\bibitem{Krizhevsky2017_Alexnet}
Krizhevsky, A., Sutskever, I., Hinton, G.E.: Imagenet classification with deep
  convolutional neural networks. Commun. ACM  \textbf{60}(6),  84–90 (May
  2017). \doi{10.1145/3065386}, \url{https://doi.org/10.1145/3065386}

\bibitem{ref_32}
Kumar, S., Mishra, S., Asthana, P., Pragya: Automated detection of acute
  leukemia using k-mean clustering algorithm. In: Bhatia, S.K., Mishra, K.K.,
  Tiwari, S., Singh, V.K. (eds.) Advances in Computer and Computational
  Sciences. pp. 655--670. Springer Singapore, Singapore (2018)

\bibitem{kusamichi04}
Kusamichi, H., Isokawa, T., Matsui, N., Ogawa, Y., Maeda, K.: {A New Scheme for
  Color Night Vision by Quaternion Neural Network}. In: Proceedings of the 2nd
  International Conference on Autonomous Robots and Agents (ICARA 2004). pp.
  101--106 (2004)

\bibitem{scotti_ALL_IDB}
Labati, R.D., Piuri, V., Scotti, F.: ALL-IDB: The acute lymphoblastic leukemia
  image database for image processing. 2011 18th IEEE International Conference
  on Image Processing  (2011). \doi{978-1-4577-1303-3}

\bibitem{mandic2009}
Mandic, D. P.;~Goh, V.S.L.: {Complex valued nonlinear adaptive filters:
  noncircularity, widely linear and neural models}, vol.~59. Wiley, New York
  (2009)

\bibitem{matsui04}
Matsui, N., Isokawa, T., Kusamichi, H., Peper, F., Nishimura, H.: {Quaternion
  neural network with geometrical operators}. Journal of Intelligent and Fuzzy
  Systems  \textbf{15}(3),  149--164 (2004)

\bibitem{MISHRA2019}
Mishra, S., Majhi, B., Sa, P.K.: Texture feature based classification on
  microscopic blood smear for acute lymphoblastic leukemia detection.
  Biomedical Signal Processing and Control  \textbf{47},  303--311 (2019).
  \doi{https://doi.org/10.1016/j.bspc.2018.08.012},

\bibitem{def}
NCI: acute lymphoblastic leukemia.
  \url{https://www.cancer.gov/publications/dictionaries/cancer-terms/def/acute-lymphoblastic-leukemia}
  (2021), accessed: 2021-02-18

\bibitem{def_linf}
NCI: lymphoblast.
  \url{https://www.cancer.gov/publications/dictionaries/cancer-terms/def/lymphoblast}
  (2021), accessed:  2021-02-18

\bibitem{metodos_diag}
NHS: Acute lymphoblastic leukaemia diagnosis.
  \url{https://www.nhs.uk/conditions/acute-lymphoblastic-leukaemia/diagnosis/}
  (2021), accessed:  2021-02-18

\bibitem{nitta2002}
Nitta, T.: {On the critical points of the complex-valued neural network}. In:
  Proceedings of the ICONIP 2002 9th International Conference on Neural
  Information Processing: Computational Intelligence for the E-Age. pp.
  411--439. Singapore (2002)

\bibitem{ogawa2016}
Ogawa, T.: {Neural network inversion for multilayer quaternion neural
  networks}. Comput Technol Appl  \textbf{7},  73--82 (2016)

\bibitem{onyekpe2021}
Onyekpe, U., Palade, V., Kanarachos, S., Christopoulos, S.R.: {A Quaternion
  Gated Recurrent Unit Neural Network for Sensor Fusion}. Information  (2021).
  \doi{10.3390/info12030117}, \url{https://doi.org/10.3390/info12030117}

\bibitem{Parcollet2020ANetworks}
Parcollet, T., Morchid, M., Linar{\`{e}}s, G.: {A survey of quaternion neural
  networks}. Artificial Intelligence Review  \textbf{53}(4),  2957--2982 (4
  2020). \doi{10.1007/s10462-019-09752-1},
  \url{https://doi.org/10.1007/s10462-019-09752-1}

\bibitem{parcollet2018}
Parcollet, T., Morchid, M., Linarès, G.: {Quaternion Convolutional Neural
  Networks for Heterogeneous Image Processing} (2018). \doi{arXiv:1811.02656v1}

\bibitem{Parcollet2018b}
Parcollet, T., Zhang, Y., Morchid, M., Trabelsi, C., Linares, G., {de Mori},
  R., Bengio, Y.: Quaternion convolutional neural networks for end-to-end
  automatic speech recognition. In: Proc. Interspeech 2018. pp. 22--26 (2018).
  \doi{10.21437/Interspeech.2018-1898},
  \url{http://dx.doi.org/10.21437/Interspeech.2018-1898}

\bibitem{Pavllo_2019}
Pavllo, D., Feichtenhofer, C., Auli, M., Grangier, D.: Modeling human motion
  with quaternion-based neural networks. International Journal of Computer
  Vision  \textbf{128}(4),  855–872 (Oct 2019).
  \doi{10.1007/s11263-019-01245-6},

\bibitem{art_historia}
Shafique, S., Tehsin, S.: Computer-aided diagnosis of acute lymphoblastic
  leukaemia. Computational and Mathematical Methods in Medicine  \textbf{2018},
   6125289 (Feb 2018). \doi{10.1155/2018/6125289},

\bibitem{shang14}
Shang, F., Hirose, A.: {Quaternion Neural-Network-Based PolSAR Land
  Classification in Poincare-Sphere-Parameter Space}. IEEE Transactions on
  Geoscience and Remote Sensing  \textbf{52},  5693--5703 (2014)

\bibitem{simonyan2015_VGG}
Simonyan, K., Zisserman, A.: Very deep convolutional networks for large-scale
  image recognition (2015)

\bibitem{art_diag}
T~Terwilliger, M.A.H.: Acute lymphoblastic leukemia: a comprehensive review and
  2017 update. Blood Cancer J.  (2017). \doi{10.1038/bcj.2017.53}

\bibitem{takahashietal2017}
Takahashi, K., Isaka, A., Fudaba, T., Hashimoto, M.: {Remarks on quaternion
  neural network-based controller trained by feedback error learning}.
  IEEE/SICE International symposium on system integration pp. 875--880 (2017)

\bibitem{takahashi_etal2014}
Takahashi, K., Takahashi, S., Cui, Y., Hashimoto, M.: Remarks on computational
  facial expression recognition from hog features using quaternion multi-layer
  neural network. In: Mladenov, V., Jayne, C., Iliadis, L. (eds.) Engineering
  Applications of Neural Networks. pp. 15--24. Springer International
  Publishing, Cham (2014). \doi{10.1007/978-3-319-11071-4\_2}

\bibitem{Trabelsi17complex}
Trabelsi, C., Bilaniuk, O., Zhang, Y., Serdyuk, D., Subramanian, S., Santos,
  J.F., Mehri, S., Rostamzadeh, N., Bengio, Y., Pal, C.J.: {Deep complex
  networks} (5 2017)

\bibitem{Tuba2019}
Tuba, M., Tuba, E.: Generative adversarial optimization (goa) for acute
  lymphocytic leukemia detection. Studies in Informatics and Control
  \textbf{28},  245--254 (10 2019). \doi{10.24846/v28i3y201901}

\bibitem{VOGADO2018}
Vogado, L.H., Veras, R.M., Araujo, F.H., Silva, R.R., Aires, K.R.: Leukemia
  diagnosis in blood slides using transfer learning in cnns and svm for
  classification. Engineering Applications of Artificial Intelligence
  \textbf{72},  415--422 (2018).
  \doi{https://doi.org/10.1016/j.engappai.2018.04.024},

\bibitem{zhu2019quaternion}
Zhu, X., Xu, Y., Xu, H., Chen, C.: Quaternion convolutional neural networks.
  In: Ferrari, V., Hebert, M., Sminchisescu, C., Weiss, Y. (eds.) Computer
  Vision -- ECCV 2018. pp. 645--661. Springer International Publishing, Cham
  (2018)

\end{thebibliography}

\end{document}